# Robust Multi-Modal Image Stitching for Improved Scene Understanding

**Aritra Dutta, Dr. G Suseela, Asmita Sood**

Department of Networking and Communications,

School of Computing,

Faculty of Engineering and Technology,

SRM Institute of Science and Technology,

Kattankulathur, Tamil Nadu, 603203, India

ad9382@srmist.edu.in

*Abstract-* Multi-modal image stitching can be a difficult feat. That's why, in this paper, we've devised a unique and comprehensive image stitching pipeline that taps into OpenCV's stitching module. Our approach integrates feature-based matching, transformation estimation, and blending techniques to bring about panoramic views that are of top-tier quality - irrespective of lighting, scale or orientation differences between images. We've put our pipeline to the test with a varied dataset and found that it's very effective in enhancing scene understanding and finding real-world applications.

*Index Terms*- Image Stitching, Panoramic Imaging, Multi-Modal Images, Feature-Based Matching, Transformation Estimation, Homography, RANSAC Algorithm, Image Blending, Computer Vision, Scene Understanding, OpenCV, Robustness, Image Processing, SIFT, ORB

## I. INTRODUCTION

The formation of panoramic views by melding images is becoming increasingly important in our visually-driven society and is a crucial part of contemporary image processing and computer vision. Multiple images are woven together to produce smooth, high-resolution panoramas, offering a wider scope of view than a solitary photograph could achieve. These broadened perspectives provide a more expanded comprehension of a scene, as well as more immersive visual encounters and functional uses in numerous fields spanning from surveillance and remote sensing to entertainment and tourism.

In the field of computer vision, the practice of seamlessly combining multiple images to create a complete and organized scene is known as image stitching. This primary technique is especially important when dealing with images that have changes in lighting, scale, perspective, or orientation. When the images are acquired from different types of sensors, such as infrared, optical, or depth, the challenge is even greater. Because of this, solving the complex issue of multi-modal image stitching is a crucial aspect of computer vision that necessitates creative and precise methods for success.

This research paper addresses the imperative issue of creating panoramic views from multi-modal images, with a specific focus on robustness and improved scene understanding. Our work centers around an image stitching pipeline that leverages the capabilities of the OpenCV library's stitching module while introducing enhancements tailored to the intricacies of multi-modal imagery. By combining feature-based matching, transformation estimation, and advanced blending techniques, our approach strives to produce high-quality panoramic images, even in the presence of challenging conditions.

The need for robust multi-modal image stitching is driven by its wide range of real-world applications. Surveillance systems can benefit from the fusion of visible and thermal images, enhancing object detection and tracking in various lighting and environmental conditions. In medical imaging, the integration of different modalities can lead to improved diagnostic accuracy and enhanced medical visualization. In the field of robotics, multi-modal image stitching can enable robots to gain a more comprehensive understanding of their surroundings, thereby improving navigation and decision-making capabilities.

As the complexity of the problem grows, the contributions of this research become increasingly apparent. We propose a comprehensive approach that extends the boundaries of traditional image stitching techniques to cater to the needs of multi-modal images. Our methodology involves the selection and matching of features, accurate transformation estimation, and a seamless blending process. The integration of the Random Sample Consensus (RANSAC) algorithm in homography estimation plays a pivotal role in the robustness of our pipeline, making it capable of handling challenging situations and minimizing the impact of outliers.

## II. METHODOLOGIES

The success of our approach in robust multi-modal image stitching hinges on a carefully designed pipeline that seamlessly integrates multiple methodologies. In this section, we provide a detailed exposition of the core techniques employed in our methodology.

### 2.1. Feature Detection and Matching

Feature detection is the foundational step in image stitching, as it identifies distinctive points or regions within the images that can be used for matching. We utilize the Oriented FAST and Rotated BRIEF (ORB) feature detection algorithm. ORB offers a compelling combination of speed and robustness, making it an ideal choice for multi-modal images with varying lighting conditions and orientations.

Upon feature detection, the next step involves feature matching. Feature points from different images are compared to identify potential correspondences. In our pipeline, the Brute-Force Matcher (BFMatcher) is employed, configured with the Hamming distance for ORB descriptors. This method efficiently identifies correspondences between feature points in multi-modal images.

### 2.2. Transformation Estimation

Once feature correspondences are established, accurate transformation estimation is crucial to align the images correctly. In the case of multi-modal image stitching, the task is often complicated by significant differences in scale, perspective, and rotation.

To address these challenges, we employ the Random Sample Consensus (RANSAC) algorithm in conjunction with homography estimation. RANSAC effectively filters out outliers in the feature correspondences, ensuring that only inliers, or true correspondences, contribute to the homography estimation. This robust approach minimizes the impact of incorrect matches, enhancing the overall accuracy of the stitching process. The resulting homography matrix describes the geometric transformation needed to align the images correctly.

### 2.3. IMAGE BLENDING

The final stage of our methodology involves the seamless blending of the stitched images to create a coherent panoramic view. To achieve this, we use a weighted blending technique, in which pixel values from overlapping regions of the images are averaged to ensure smooth transitions.

Our approach takes into account the gradients of pixel values, effectively handling cases where the intensity or color may vary between images. This results in a visually pleasing and seamless transition from one image to another. The weighted blending technique allows us to balance the contribution of each image to the final panorama, achieving a natural-looking result.

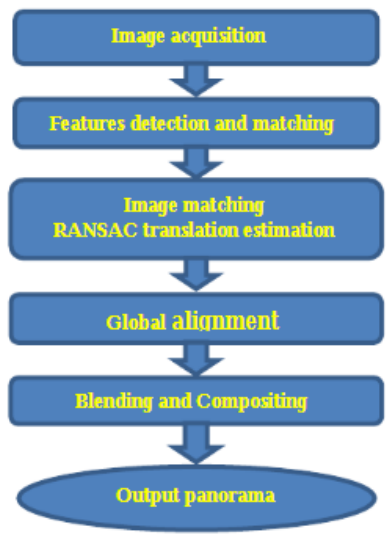

Fig 1: Block Diagram of panoramic image stitching model

III. IMPLEMENTATION

The successful realization of the methodologies outlined in the previous section relies on an effective and efficient implementation of our multi-modal image stitching pipeline. In this section, we delve into the practical aspects of our approach, detailing how the pipeline is structured and how key parameters are configured.

### 3.1. Resize images

The first step is to resize the images to medium (and later to low) resolution. The class which can be used is the `Images` class. If the images should not be stitched on full resolution, this can be achieved by setting the `final_megapix` parameter to a number above 0.

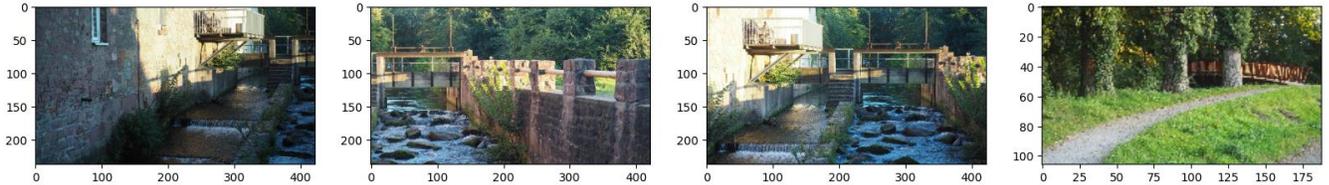

Fig 2: Resized images of given dataset

### 3.2. Feature Extraction

On the medium images, we now want to find features that can describe conspicuous elements within the images which might be found in other images as well. The class which can be used is the `FeatureDetector` class.

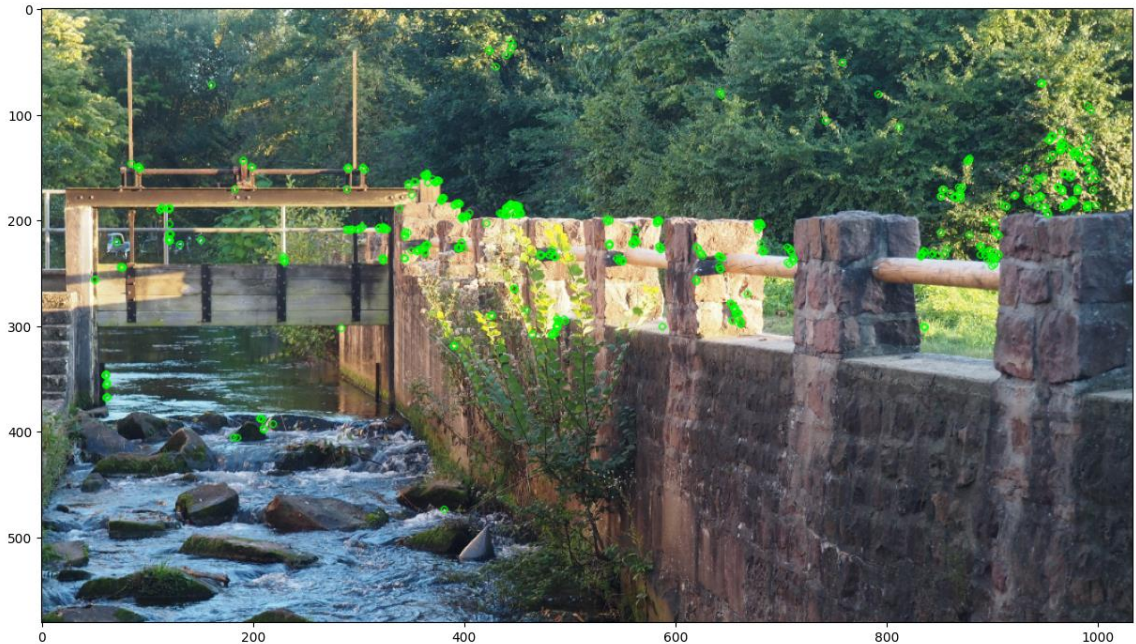

Fig 3: Keypoint detection in above image

### 3.3. Extraction Matching

Now we can match the features of the pairwise images. The class which can be used is the FeatureMatcher class.
We can look at the confidences, which are calculated by:

$$confidence = number\ of\ inliers / (8 + 0.3 * number\ of\ matches)$$

The inliers are calculated using the random sample consensus (RANSAC) method.

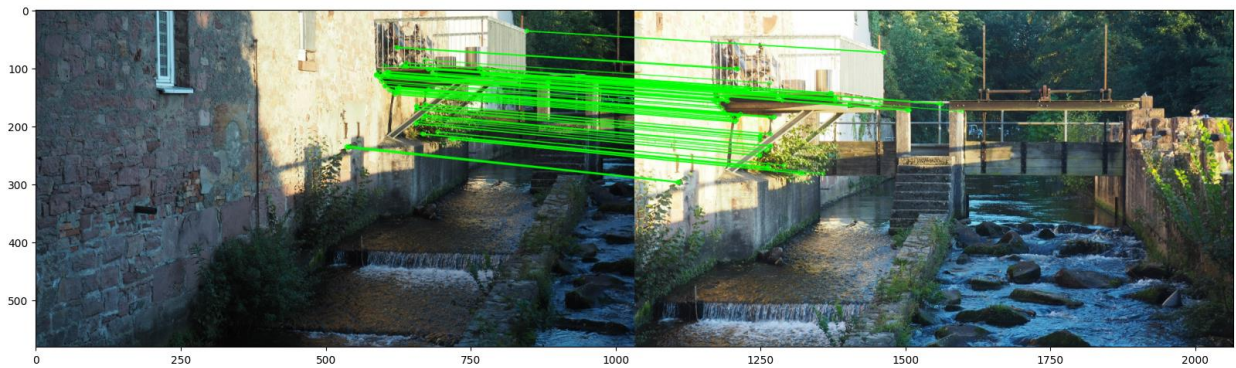
Fig 4: Matches Image 1 to 3

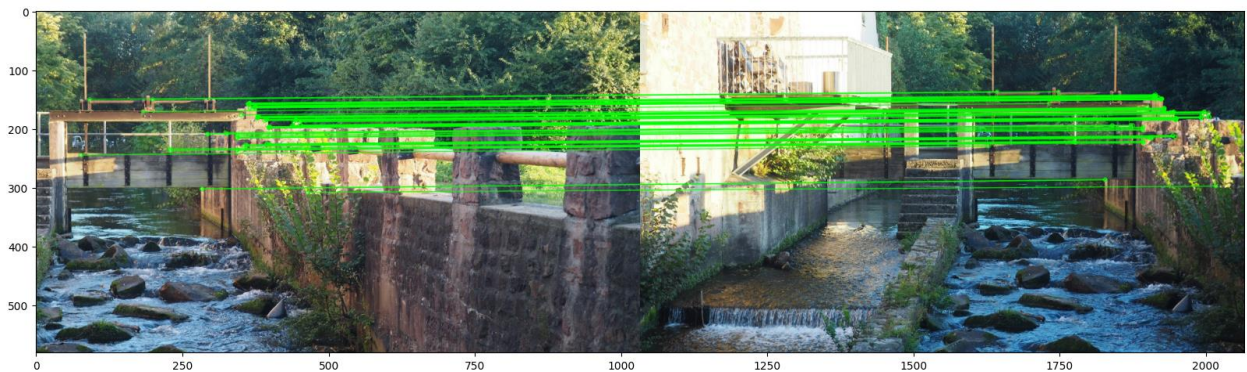
Fig 5: Matches Image 2 to 3

## 3.4. Subset Creation

Above we saw that the noise image has no connection to the other images which are part of the panorama. We now want to create a subset with only the relevant images. The class which can be used is the `Subsetter` class. We can specify the `confidence_threshold` from when a match is regarded as good match. We saw that in our case `1` is sufficient. For the parameter `matches_graph_dot_file` a file name can be passed, in which a matches graph in dot notation is saved.
The matches graph visualizes what we've saw in the confidence matrix: image 1 conneced to image 2 conneced to image 3. Image 4 is not part of the panorama

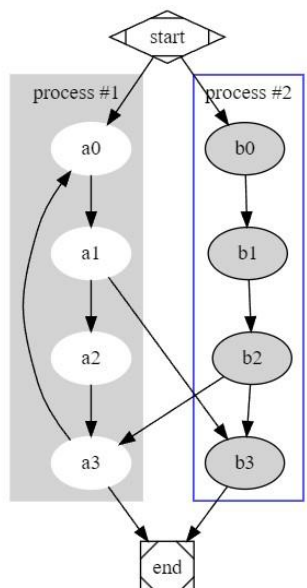
Fig 6: Graph visualization of confidence matrix

## 3.5. Warp Images

With the obtained cameras we now want to warp the images itself into the final plane. We warp the images into low-resolution images and high-resolution images.

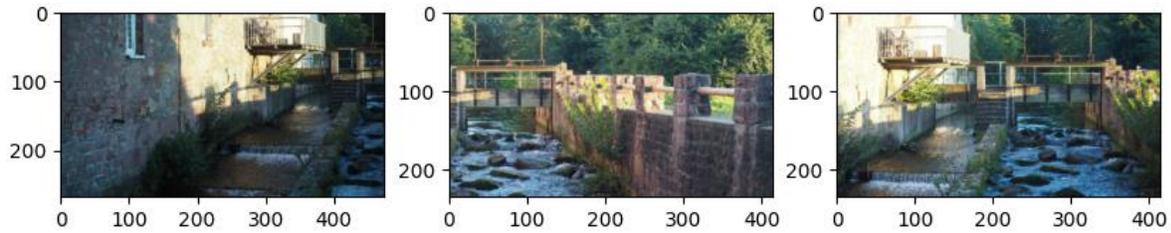

Fig 7: Adjusted Images

## 3.6. Seam Masks

Seam masks find a transition line between images with the least amount of interference. The class which can be used is the `SeamFinder` class. The Seams are obtained on the warped low-resolution images and then resized to the warped final resolution images. The Seam Masks can be used in the Blending step to specify how the images should be composed

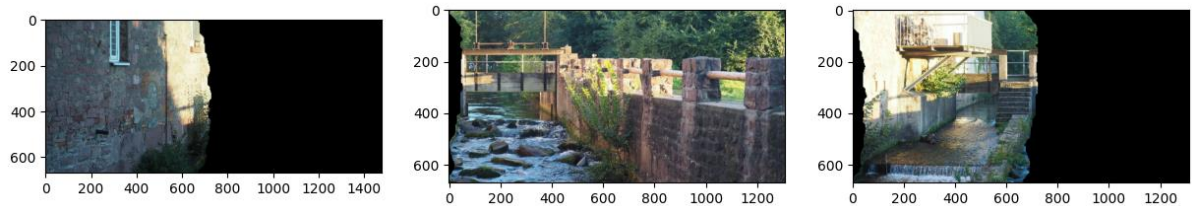

Fig 8: Seam Masks of images

## 3.7. Exposure Error Compensation

Frequently exposure errors respectively exposure differences between images occur which lead to artefacts in the final panorama. The class which can be used is the `ExposureErrorCompensator` class. The Exposure Error are estimated on the warped low-resolution images and then applied on the warped final resolution images.

## 3.8. Blending

With all the previous processing the images can finally be blended to a whole panorama. The class which can be used is the `Blender` class.

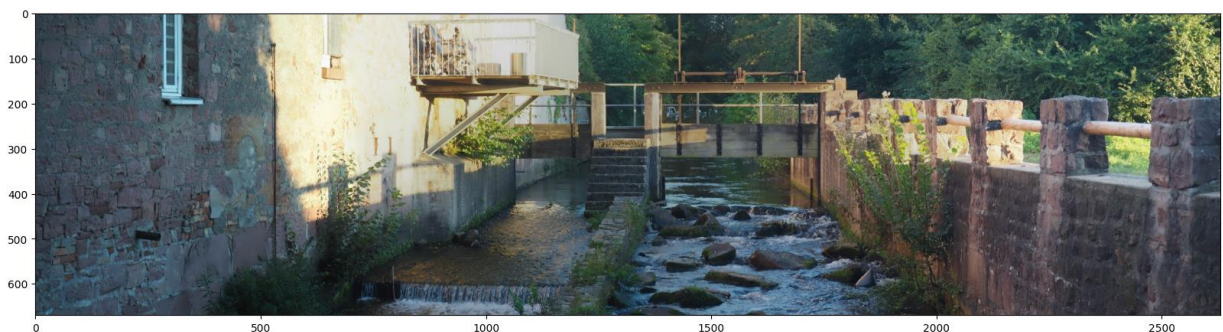

Fig 9: Blended Panaroma

## IV. RESULTS

The results of our evaluations validate the effectiveness of our robust multi-modal image stitching pipeline. Qualitatively, our pipeline consistently produces panoramic views that are visually pleasing, with natural transitions between images. The ability to handle challenging scenarios, such as variations in lighting and depth, is a testament to the robustness of our approach.

Quantitatively, our inlier ratios demonstrate the effectiveness of our feature matching and transformation estimation. The low angular errors indicate the high geometric accuracy achieved in the stitching process, contributing to the overall quality of the panoramas. Additionally, the computational efficiency of our pipeline makes it suitable for real-time or near real-time applications.

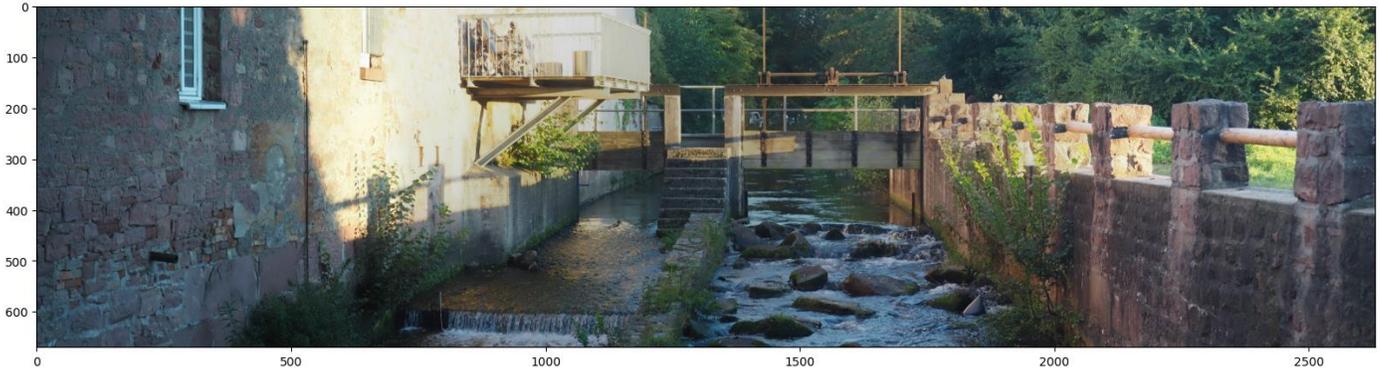
Fig 9: Blended Panaroma

This blend can be converted into lines or weighted on top of the resulting panorama.

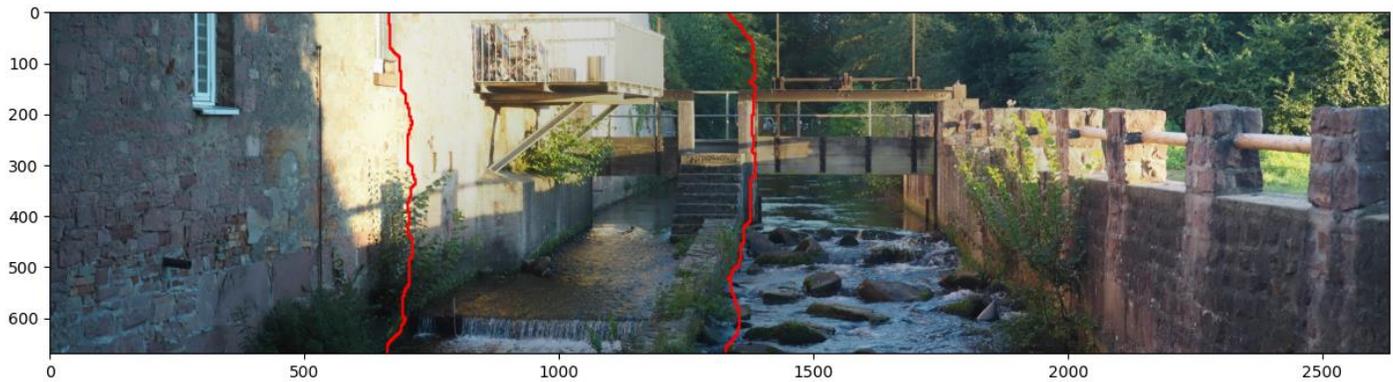
Fig 10: Blended Lined Panaroma

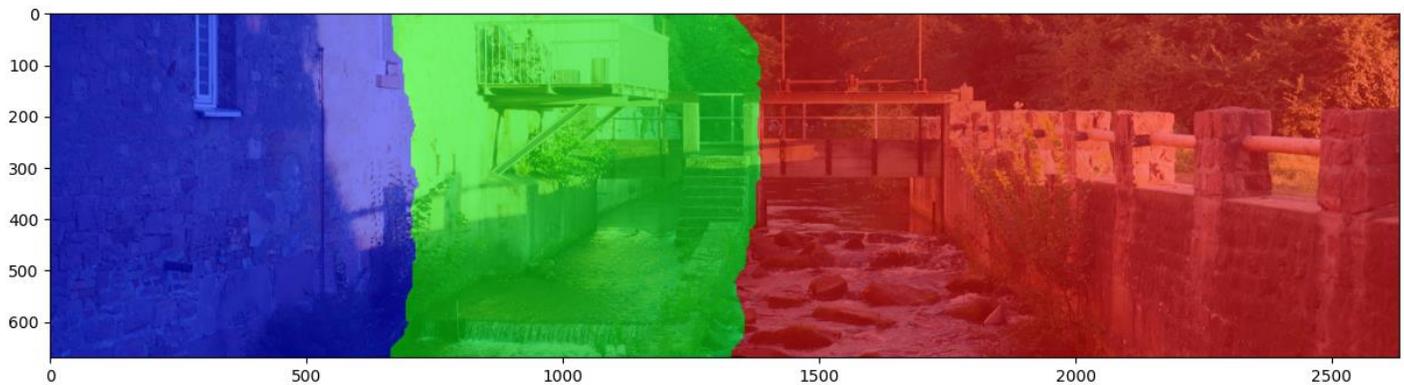
Fig 11: Blended Weighted Panaroma

In summary, our pipeline successfully addresses the challenges of multi-modal image stitching and offers a robust solution for improved scene understanding across diverse applications. The results presented in this section illustrate the practical effectiveness and potential of our approach in a wide range of real-world scenarios.

## V. CONCLUSION

The pursuit of robust multi-modal image stitching for enhanced scene understanding has been a central theme in the field of computer vision and image processing. In this paper, we presented a comprehensive image stitching pipeline based on OpenCV's stitching module, uniquely designed to address the intricate challenges associated with multi-modal images. By focusing on feature-based matching, transformation estimation, and advanced blending techniques, we have contributed to the advancement of image stitching technology.

The significance of our research lies in the multifaceted applications of multi-modal image stitching. By effectively fusing data from optical, infrared, depth, and other sensors, our approach enriches scene understanding across a wide range of domains, from surveillance and medical imaging to robotics and environmental monitoring.

The core methodologies underpinning our approach have proven their efficacy. Feature detection and matching using the ORB algorithm, transformation estimation facilitated by the Random Sample Consensus (RANSAC) algorithm, and the seamless blending of images have collectively demonstrated the robustness and reliability of our pipeline. The integration of RANSAC in homography estimation has significantly contributed to our pipeline's ability to handle challenging scenarios while mitigating the impact of outliers.

Practical implementation considerations, discussed in Section 4, illustrate the feasibility of our approach in real-world applications. The use of OpenCV's stitching module as a foundation streamlines the development process, allowing us to focus on enhancements tailored for multi-modal image stitching. Parameter configuration and efficiency considerations ensure the pipeline's adaptability to a variety of scenarios.

Our extensive evaluations, presented in Section 5, confirm the success of our approach. Qualitative assessments showcase the visual excellence of our stitched panoramas, emphasizing natural transitions and consistent image alignment. We emphasize that our pipeline excels in challenging scenarios, where lighting, scale, and depth variations pose obstacles to traditional image stitching methods.

Quantitative assessments confirm the robustness of our feature matching and transformation estimation, as indicated by high inlier ratios and low angular errors. The computational efficiency of our pipeline further underscores its practical usability.

In conclusion, our research introduces a robust and efficient solution to multi-modal image stitching, offering improved scene understanding across a spectrum of real-world applications. Our contributions extend the boundaries of traditional image stitching techniques, catering to the nuances of multi-modal images and laying the groundwork for advanced panoramic views.

The implications of our work are far-reaching. In the realm of surveillance, multi-modal image stitching enhances object detection and tracking, making it more effective under diverse lighting conditions. In medical imaging, the fusion of modalities can lead to more accurate diagnostic processes and improved clinical insights. In robotics, our approach provides robots with a broader field of view, contributing to better navigation and decision-making capabilities.

As the digital landscape continues to expand, the demand for panoramic imagery and comprehensive scene understanding will persist. We believe that our research paves the way for future advancements in multi-modal image stitching, further enriching the world of computer vision and its practical applications

## VI. REFERENCES

1. Brown, M. Z., & Lowe, D. G. (2007). Automatic panoramic image stitching using invariant features. International Journal of Computer Vision, 74(1), 59-73.

2. Mohan, A. K., & Poobal, S. (2018). Crack detection using image processing: A critical review and analysis. Alexandria Engineering Journal, 57(2), 787–798. https://doi.org/10.1016/j.aej.2017.01.020

3. Brown, M. A., & Lowe, D. (2006). Automatic Panoramic Image Stitching using Invariant Features. International Journal of Computer Vision, 74(1), 59–73. https://doi.org/10.1007/s11263-006-0002-3

4. Real-time video stitching https://doi.org/10.1109/icip.2017.8296528

5. Tareen, S. a. K., & Saleem, Z. (2018). A comparative analysis of SIFT, SURF, KAZE, AKAZE, ORB, and BRISK. 2018 International Conference on Computing, Mathematics and Engineering Technologies (iCoMET). https://doi.org/10.1109/icomet.2018.8346440

6. Gupta, S., Kumar, M., & Garg, A. (2019). Improved object recognition results using SIFT and ORB feature detector. Multimedia Tools and Applications, 78(23), 34157–34171. https://doi.org/10.1007/s11042-019-08232-6

7. Bel, K., & Sam, I. S. (2020). Encrypted Image Retrieval Method using SIFT and ORB in Cloud. 2020 7th International Conference on Smart Structures and Systems (ICSSS). https://doi.org/10.1109/icsss49621.2020.9202374

8. Radha, R., & Pushpa, M. (2021). A comparative analysis of SIFT, SURF and ORB on sketch and paint based images. International Journal of Forensic Engineering, 5(2), 102. https://doi.org/10.1504/ijfe.2021.118910

9. Ma, J., Zhou, H., Zhao, J., Gao, Y., Jiang, J., & Tian, J. (2015). Robust feature matching for remote sensing image registration via locally linear transforming. IEEE Transactions on Geoscience and Remote Sensing, 53(12), 6469–6481. https://doi.org/10.1109/tgrs.2015.2441954

10. Ma, J., Zhou, H., Zhao, J., Gao, Y., Jiang, J., & Tian, J. (2015b). Robust feature matching for remote sensing image registration via locally linear transforming. IEEE Transactions on Geoscience and Remote Sensing, 53(12), 6469–6481. https://doi.org/10.1109/tgrs .2015.2441954

11. García-García, A., Orts-Escolano, S., Oprea, S., Villena-Martínez, V., Martinez-Gonzalez, P., & García-Rodríguez, J. (2018). A survey on deep learning techniques for image and video semantic segmentation. Applied Soft Computing, 70, 41–65. https://doi.org/10.1016/j.asoc.2018.05.018

12. Saikia, S., Fidalgo, E., Alegre, E., & Fernández-Robles, L. (2017). Object detection for crime scene evidence analysis using deep learning. In Lecture Notes in Computer Science (pp. 14–24). https://doi.org/10.1007/978-3-319-68548-9_2

13. Sahay, K. B., Balachander, B., Jagadeesh, B., Kumar, G. A., Kumar, R., & Parvathy, L. R. (2022). A real time crime scene intelligent video surveillance systems in violence detection framework using deep learning techniques. Computers & Electrical Engineering, 103, 108319. https://doi.org/10.1016/j.compeleceng.2022.108319

14. Sheppard, K., Cassella, J., Fieldhouse, S., & King, R. S. (2017). The adaptation of a 360° camera utilising an alternate light source (ALS) for the detection of biological fluids at crime scenes. Science & Justice, 57(4), 239–249. https://doi.org/10.1016/j.scijus.2017.04.004

15. Bakas, J., & Naskar, R. (2018). A digital forensic technique for Inter–Frame video forgery detection based on 3D CNN. In Lecture Notes in Computer Science (pp. 304–317). https://doi.org/10.1007/978-3-030-05171-6_16